\documentclass{article}

\usepackage[final, nonatbib]{neurips_2020_ml4ps}

\usepackage[utf8]{inputenc} 
\usepackage[T1]{fontenc}    
\usepackage{hyperref}       
\usepackage{url}            
\usepackage{booktabs}       
\usepackage{amsfonts}       
\usepackage{nicefrac}       
\usepackage{microtype}      
\usepackage{graphicx}
\usepackage{amsmath}
\usepackage{xcolor}
\usepackage{algorithm2e}
\usepackage{float}

\definecolor{C0}{HTML}{1F77B4}
\definecolor{C1}{HTML}{FF7F0E}
\definecolor{C2}{HTML}{2CA02C}
\definecolor{C3}{HTML}{D62728}

\usepackage{todonotes}
\definecolor{darkspringgreen}{rgb}{0.09, 0.45, 0.27}

\title{Learning summary features of time series for likelihood free inference}

\author{%
  Pedro L. C. Rodrigues, Alexandre Gramfort \\
  Parietal team, Inria, Palaiseau, France \\
  \texttt{pedro.rodrigues@inria.fr}
}

\begin{document}

\maketitle

\begin{abstract}
There has been an increasing interest from the scientific community in using \emph{likelihood-free} inference (LFI) to determine which parameters of a given simulator model could best describe a set of experimental data. Despite exciting recent results and a wide range of possible applications, an important bottleneck of LFI when applied to time series data is the necessity of defining a set of summary features, often hand-tailored based on domain knowledge. In this work, we present a data-driven strategy for automatically learning summary features from univariate time series and apply it to signals generated from autoregressive--moving-average (ARMA) models and the Van der Pol Oscillator. Our results indicate that learning summary features from data can compete and even outperform LFI methods based on hand-crafted values such as autocorrelation coefficients even in the linear case.
\end{abstract}

\section{Introduction}
\label{sec:intro}
Bayesian inference on modern complex simulator models is in general a difficult task because the likelihood function of the model's output is often difficult or impossible to obtain. A modern approach for bypassing such an obstacle is to use \emph{likelihood-free} inference methods, in which a flexible function (e.g. a neural network) learns how to approximate the posterior distribution based on simulations over different parameters. This approach is also called \emph{simulation-based inference} (SBI) and we refer the reader to~\cite{Cranmer201912789} for a review on the topic and the Python package \texttt{sbi} for ready-to-use implementations~\cite{tejero-cantero2020sbi}. 

A common step in SBI is to reduce the dimensionality of the observed data before approximating the posterior distribution. This is particularly relevant when working with time series data, for which descriptions based on a few meaningful parameters is standard practice. For instance, the autocorrelation function is a sufficient statistic for signals generated by Gaussian autoregressive--moving-average (ARMA) models and is, therefore, the ideal summary feature extractor. For signals generated by complex non-linear simulators, such as the Hodgkin-Huxley or Lotka-Volterra models, the summary features are usually chosen based on previous domain knowledge~\cite{papamakarios19a}. In this work, we propose a data-driven strategy that avoids tailoring summary statistics for each application.

Automatically learning summary features is a recurring theme in the SBI literature. Fearnhead et al.~\cite{Fearnhead2010} present a regression scheme that approximates the posterior mean via a linear model and uses it as summary features for learning the posterior distribution. Partially exchangeable networks (PEN)~\cite{Wiqvist2019} explores invariances in time series to learn a set of summary features with the same loss function as~\cite{Fearnhead2010} and then approximates the posterior distribution using them. An important difference of our work is that we learn the summary statistics and the posterior distribution jointly, as done in~\cite{Greenberg2019} but with a different summary feature extractor.

In what follows, we present the architecture of our summary feature extractor and show its performance in approximating posterior distributions of linear and non-linear stochastic time series models. 

\section{Methods}
\label{sec:methods}
The basic setup of Bayesian methods for SBI consists of a prior distribution $p(\boldsymbol{\theta})$ over the parameters and a stochastic simulator whose output for a given $\boldsymbol{\theta}_i$ is $\boldsymbol{x}_i \sim p(\boldsymbol{x}|\boldsymbol{\theta}_i)$. Our goal is to determine the posterior distribution of $\boldsymbol{\theta}$ given an observed simulator output $\boldsymbol{x}_0 \in \mathbb{R}^{n_s}$, which we denote $p(\boldsymbol{\theta}|\boldsymbol{x}_0)$.

\subsection{Posterior estimation with adaptive summary features}
We define the summary feature extractor as a function $f_{\boldsymbol{\lambda}}: \mathbb{R}^{n_s} \to \mathbb{R}^{n_f}$ whose parameters $\boldsymbol{\lambda}$ can be learned from the data and use a probability density estimator $q_{\boldsymbol{\phi}}$ parametrized by $\boldsymbol{\phi}$ to approximate $p(\boldsymbol{\theta}|\boldsymbol{x})$. We have, therefore, that $p(\boldsymbol{\theta}|\boldsymbol{x}_0) \approx q_{\boldsymbol{\phi}}(\boldsymbol{\theta}|f_{\boldsymbol{\lambda}}(\boldsymbol{x}_0))$.

The parameters $\boldsymbol{\phi}$ and $\boldsymbol{\lambda}$ are determined by jointly minimizing the Kullback-Leiber divergence between the true posterior distribution and our approximation, which can be rewritten as 
\begin{equation}
\label{eq:loss_function}
\mathcal{L}(\boldsymbol{\phi}, \boldsymbol{\lambda}) = \mathbb{E}_{(\boldsymbol{x}, \boldsymbol{\theta}) \sim p(\boldsymbol{x}, \boldsymbol{\theta})}\Big[-\log(q_{\boldsymbol{\phi}}(\boldsymbol{\theta}|f_{\boldsymbol{\lambda}}(\boldsymbol{x})))\Big]~.
\end{equation}
The standard approach for minimizing~\eqref{eq:loss_function} is to generate a set of paired samples $(\boldsymbol{\theta}_i, \boldsymbol{x}_i) \sim p(\boldsymbol{\theta}, \boldsymbol{x})$ and use stochastic gradient descent to obtain a minimizer. However, we have observed superior results with sequential neural posterior estimation (SNPE-C)~\cite{Greenberg2019}. It consists of an $R$-step procedure in which $N$ paired samples are generated per iteration and used to obtain a sequence of approximations of $p(\boldsymbol{\theta}|\boldsymbol{x})$. SNPE-C's particularity is that the data points generated on a given round $r$ are sampled from a posterior approximation obtained on round $r-1$ fixed with $\boldsymbol{x} = \boldsymbol{x}_0$. This leads to approximations which are targeted to better represent the posterior distribution associated with the observed data point. As a result, the approximation of $p(\boldsymbol{\theta}|\boldsymbol{x}_0)$ after $R$ rounds is often better than with a single round and $N \times R$ simulations. 

{In all experiments below, $q_{\boldsymbol{\phi}}$ is a normalizing flow~\cite{papamakarios17a} with an autoregressive architecture implemented via the masked autoencoder for distribution estimation (MADE)~\cite{germain15}. We follow the same setup from~\cite{Durkan2020} and~\cite{Greenberg2019} for LFI problems, stackings five MADEs, each with two hidden layers of 50 units, and a standard normal base distribution for the normalizing flow. This choice provides a sufficiently flexible function capable of approximating complex posterior distributions. We refer the reader to~\cite{papamakarios2019normalizing} for more information on the different types of normalizing flows. We run SNPE-C with $R = 10$ rounds and $N = 5000$ simulations per round.}

\subsection{The \textit{YuleNet}}
The summary feature extractor that we propose is based on a previously published neural network architecture presented in~\cite{Chambon2018}. For an input time series realization $\boldsymbol{x} \in \mathbb{R}^{n_s}$ ($n_s > 256$), {we get an $n_f$-dimensional feature vector given by}
\begin{equation}
f_{\boldsymbol{\lambda}}(\boldsymbol{x}) = g_3\big(g_2\big(g_1(\boldsymbol{x})\big)\big) \in \mathbb{R}^{n_f}~,
\end{equation} 
where $g_{1}, g_{2}, $ and $g_{3}$ are functions with trainable parameters,
\begin{itemize}
  \setlength\itemsep{0em}
  \item[$g_1$~:] 1D CNN (input channels: 1, output channels: 8, kernel size: 64, stride: 1, padding: 32) followed by a ReLU activation function and an average pooling (kernel size: 16)

  \item[$g_2$~:] 1D CNN (input channels: 8, output channels: 8, kernel size: 64, stride: 1, padding: 32) followed by a ReLU activation function and an average pooling (kernel size: $\lfloor n_s / 256 \rfloor$)

  \item[$g_3$~:] 50\%-dropout layer followed by a linear transformation with $n_f$-dimensional output and a ReLU activation function.
\end{itemize}
We name our summary feature extractor \textit{YuleNet} in reference to the widely known Yule-Walker equations from time series analysis~\cite{durbin1960fitting}. A \textit{YuleNet} contains $4624 + n_f(1+n_s/32)$ trainable parameters in total. In all of the examples considered in this paper, we have $n_s = 4096$ and $n_f = 5$, leading to 5269 parameters to optimize.

\subsection{Examples on linear time series processes}
We consider two examples on discrete-time series generated by stochastic linear models. Despite their simplicity, these models are good benchmarks for validating our approach since the posterior distributions are known analytically.

Our first example is an auto-regressive process of order two parametrized by $k_1, k_2 \in (-1, +1)$, 
\begin{equation}
  \label{eq:AR-process}
  x(n) = (k_1 + k_1 k_2) x(n-1) + |k_2| x(n-2) + u(n)~,
\end{equation} 
where $u$ is a zero-mean discrete-time Gaussian white noise process with unit variance. Note that the absolute value taken over $k_2$ leads to a posterior distribution $\boldsymbol{\theta} = (k_1, k_2)$ which has two modes. We call this example \texttt{bimodal-AR(2)}. Our second example is a moving-average process of order two, \texttt{MA(2)}:
\begin{equation}
\label{eq:MA-process}
x(n) = (k_1 + k_1 k_2) \epsilon(n-1) + k_2 \epsilon(n-2) + u(n)~, 
\end{equation}
with $k_1, k_2,$ and $u$ defined as in \eqref{eq:AR-process}. Note that the parametrization in terms of $k_1, k_2$ ensures that \texttt{bimodal-AR(2)} is a stationary process and that \texttt{MA(2)} is an invertible process~\cite{boxjen76}. 

In both examples, we use SNPE-C to approximate the posterior distribution $p(\boldsymbol{\theta}|\boldsymbol{x}_0)$ with $\boldsymbol{x}_0$ generated by $\boldsymbol{\theta}_0 = (k_1, k_2) = (0.5, -0.75)$. We consider an uniform prior distribution for the parameters $k_1, k_2 \sim \mathcal{U}(-1, +1)$.

\subsection{The Van der Pol oscillator}
The Van der Pol oscillator (\texttt{VdPOsc}) is a non-linear system widely known in the study of chaotic systems. It has been used to decribe different phenomena in physics, biology, and electrical engineering~\cite{strogatz2000}. In this example, we consider a stochastic version driven by a random process~\cite{Belousov2020}. For a continuous time series $x(t)$, we have
\begin{equation}
\label{eq:VanDerPol}
\ddot{x} = \varepsilon (1-x^2) \dot{x} - x + \sigma\dot{w}~,
\end{equation}
where $\varepsilon > 0$ and $\sigma > 0$, and $\dot{w}(t)$ is a Gaussian white noise of zero mean and unit variance. 

We solve~\eqref{eq:VanDerPol} using a Euler-Maruyama solver for Ito equations with $\Delta t = 0.01s$ and then downsample the time series to a sampling period of $T_s = 0.05s$. The initial conditions are fixed to $(x(0), \dot{x}(0)) = (1.0, 2.0)$. We use SNPE-C to approximate the posterior distribution $p(\boldsymbol{\theta}|\boldsymbol{x}^{(i)}_0)$ with ${\boldsymbol{\theta}} = (\varepsilon, \sigma)$ in five different cases:
\begin{equation*}
\begin{array}{ccccc}
\boldsymbol{\theta}^{(1)}_0 = (2.5, 1.0) &
\boldsymbol{\theta}^{(3)}_0 = (1.0, 1.5) & \boldsymbol{\theta}^{(5)}_0 = (4.0, 0.5) &
\boldsymbol{\theta}^{(2)}_0 = (1.0, 0.5) & \boldsymbol{\theta}^{(4)}_0 = (4.0, 1.5)
\end{array}
\end{equation*}
We consider a prior distribution for $\varepsilon \sim \mathcal{U}(0, 5)$ and $\sigma \sim \mathcal{U}(0, 2)$.

\section{Results and discussion}
\label{sec:results}

{\color{black}\textbf{Examples with linear time series processes.} We assess the quality of the results with SNPE-C via the Wasserstein distance~\cite{Peyre2019} calculated over samples from the analytic and approximated posteriors of \texttt{bimodal-AR(2)} and \texttt{MA(2)}. We compare the results obtained when the summary features are extracted via the {\color{C1}YuleNet} architecture to when they are estimates of the autocorrelations of the time series ({\color{C0} autocorr}). Note that because $u$ is a Gaussian white noise in both~\eqref{eq:AR-process} and~\eqref{eq:MA-process}, the autocorrelations are sufficient statistics, making them the ideal benchmark. We also consider a case where $f_{\boldsymbol{\lambda}}$ is a PEN with the same setup from~\cite{Wiqvist2019} in two situations: when its parameters are learned within the SNPE-C procedure ({\color{C2}PEN}), and when they are learned separately on $10^4$ simulated samples and then applied to a rejection-sample ABC procedure ({\color{C3}PEN+ABC}) as done in~\cite{Wiqvist2019}.} 

\begin{figure}
  \centering
  \begin{minipage}{\linewidth}
      \begin{minipage}{0.65\linewidth}
          \includegraphics[width=\textwidth]{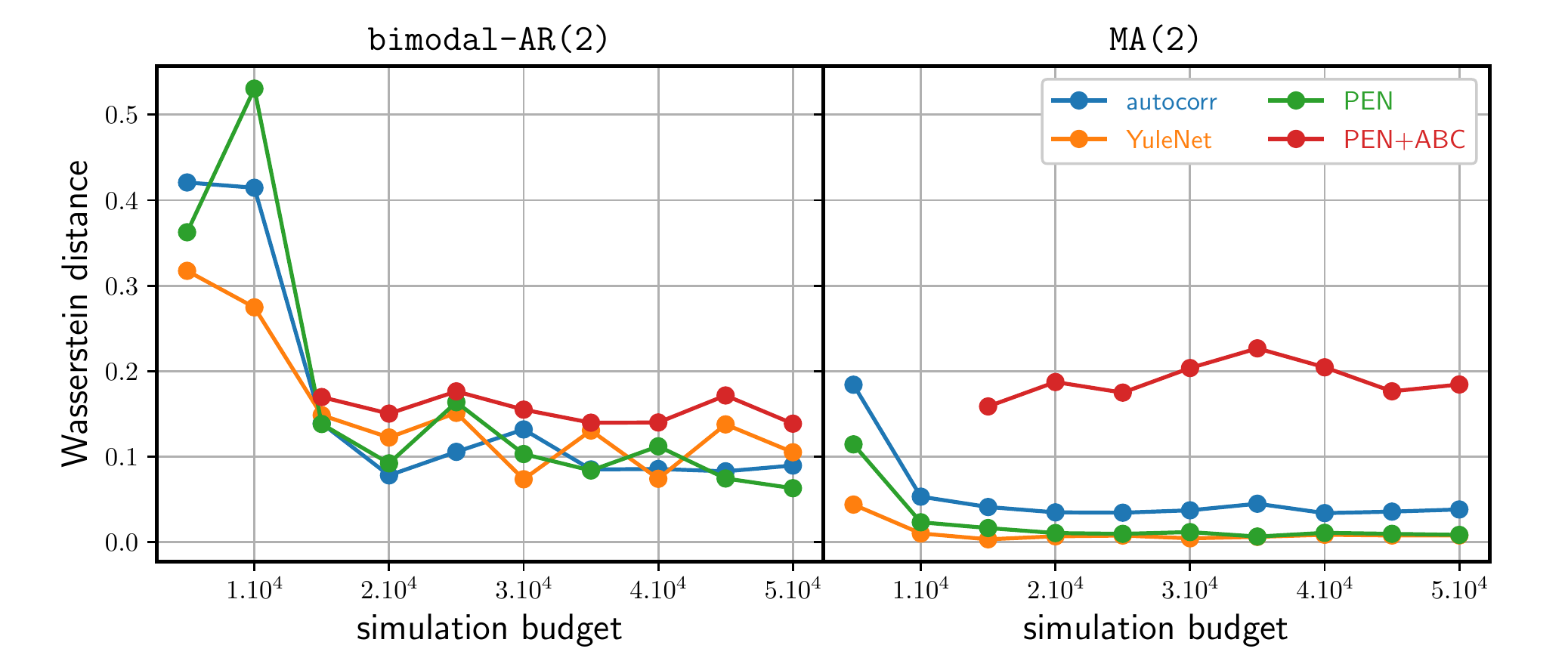}
      \end{minipage}
      \begin{minipage}{0.35\linewidth}
  \caption{\label{fig:results-arma}Wasserstein distance between the estimated posterior and the analytical posterior distribution for the examples on \textbf{linear time series processes}. The distances have been averaged over 100 batches with 100 samples from each posterior distribution (all calculations done with the \texttt{POT} package~\cite{flamary2017pot})}
       \end{minipage}
  \end{minipage}
\end{figure}

Figure~\ref{fig:results-arma} shows that the Wasserstein distances of all methods decrease when the simulation budget increases (simulation budget = round number $\times$ simulations per round). In \texttt{bimodal-AR(2)}, the performances of {\color{C0} autocorr}, {\color{C1}YuleNet}, and {\color{C2}PEN} are very similar, whereas {\color{C3}PEN+ABC} is slightly worse. It is worth mentioning that {\color{C2}PEN} has twice more parameters as compared to {\color{C1}YuleNet} (10285 versus 5269) and requires $\approx 27.10^6$ multiply-accumulate operations (MACs) per forward computation, against $\approx 3.10^6$ for {\color{C1}YuleNet}. {In practice, we have observed that the evaluations of {\color{C2}PEN}s are between three and five times longer in CPU time than with {\color{C1}YuleNet}}. In \texttt{MA(2)}, we observe that {\color{C1}YuleNet} and {\color{C2}PEN} are uniformly better than {\color{C0}autocorr}, indicating that the learned summary features in this case are better representations of the time series for the inference procedure than the estimated autocorrelations. This can be explained by the highly complex relation between the theoretical values of the autocorrelations of \texttt{MA(2)} and parameters $k_1$ and $k_2$, which makes the model inversion via SNPE-C very difficult (see Appendix~\ref{appendix} for details).

\textbf{Results on the Van der Pol oscillator.} We are not aware of any direct method for obtaining samples of the true posterior distribution of \texttt{VdPOsc}. Therefore, we base our discussion on the inspection of 2D histograms of the samples from each approximation to $p(\boldsymbol{\theta}|\boldsymbol{x}_0)$. We consider results only for {\color{C0}autocorr} and {\color{C1}YuleNet}.

\begin{figure}
  \centering
  \includegraphics[width=\textwidth]{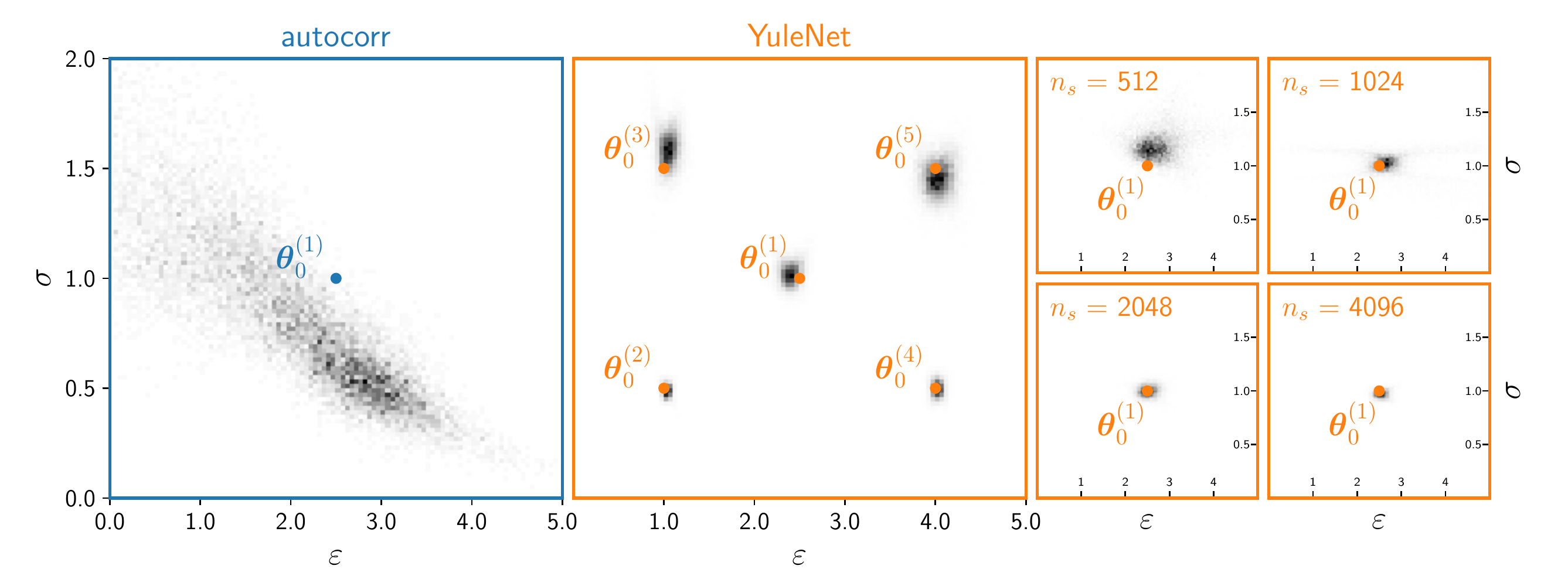}
  \caption{\label{fig:results-vanderpol}Results on the \textbf{Van der Pol oscillator} after ten SNPE-C rounds. The {left} plot portrays a 2D histogram of $10^4$ samples from the posterior distribution approximated with {\color{C0}autocorr} for ground truth parameter $\boldsymbol{\theta}^{(1)}_0$. It shows that autocorrelations are not adequate summary statistics for this example. The center plot shares the y-axis of the left plot and overlays the histograms from approximations estimated with {\color{C1}YuleNet} on five different ground truth parameters ($10^4$ samples for each case). We see how the shape of the posterior distribution varies for different choices of ground truth parameter $\boldsymbol{\theta}^{(i)}_0$. The four small plots on the right demonstrate that the dispersion of the samples with {\color{C1}YuleNet} decrease when longer time series are used in the inference procedure.}
\end{figure}

The left and center plots of Figure~\ref{fig:results-vanderpol} show that the posterior distribution approximated with {\color{C0}autocorr} generates samples which are very far from the ground truth parameter, whereas {\color{C1}YuleNet} yields sharp posterior distributions around the true parameter. This indicates that using only autocorrelations as summary features for a non-linear model such as \texttt{VdPOsc} is not sufficient to capture its statistical properties and that the flexibility of {\color{C1}YuleNet} allows for a better representation of the time series. We also observe that the dispersions of the samples of the posterior distributions obtained for {\color{C1}YuleNet} vary for different ground truth parameters: the standard deviation $\sigma$ of the input Gaussian white noise has the usual effect of increasing the variance of the posterior distribution, and the `degree' of non-linearity of \texttt{VdPOsc}, modulated by $\varepsilon$, shows that when the system tends to an harmonic oscillator ($\varepsilon \to 0$) the posterior distribution has less variance. The four small plots on the right side of Figure~\ref{fig:results-vanderpol} display how the dispersion of the samples from the posterior approximation vary for different lengths of the observed time series. They show that our approximate posterior has the expected behavior of getting sharper when longer observations are available.


\section{Conclusion}
\label{sec:conclusion}

Our experiments have demonstrated that data-driven summary features of time series learnt for SBI can match the performance of sufficient statistics in the Gaussian linear case, while being significantly better on a non-linear stochastic model. The {\color{C1}YuleNet} architecture that we present, which is simply a well calibrated one-dimensional CNN, learns adequate summary features for inference while using less parameters and requiring less computational power (measured via MACs) to train than another proposal from the literature. Using the SNPE-C procedure to minimize~\eqref{eq:loss_function} and a normalizing flow for $q_{\boldsymbol{\phi}}$ has given satisfactory results, but other options are possible too, such as using MDNs~\cite{Bishop94mixturedensity} to approximate probability distributions and employing other LFI strategies as in~\cite{papamakarios19a} and~\cite{hermans2020likelihoodfree}.

\section*{Broader Impact}
Inferring the laws and parameters that drive physical systems has been a long standing issue in physics, and more broadly across all experimental sciences.
In this work, we have presented a method for automatically determining summary statistics from experimental time series and used them with a LFI method to obtain the parameters of a simulator model. Our main contribution is in easing the burden on specialists interested in applying LFI in practice, since the tailoring of summary statistics for each application is often time consuming. 

%
%

\bibliographystyle{plain}
\bibliography{biblio.bib}

\medskip

\small

\section{Appendix}
\label{appendix}
The auto-correlation function for bimodal-AR(2) in terms of $(k_1, k_2)$ is
\begin{equation}
r(1) = \frac{k_1+k_1 |k_2|}{1 - |k_2|}~,~r(2) = |k_2| + \frac{(k_1 + k_1 |k_2|)^2}{1 - |k_2|}~,
\end{equation}
and $r(s) = a_1 r(s-1) + a_2 r(s-2), \forall s > 2, s \in \mathbb{Z}$. The auto-correlation function for MA(2) is
\begin{equation}
\quad r(1) = \frac{(k_1+k_1 k_2)(1+k_2)}{1+(k_1 + k_1 k_2)^2+k_2^2} \quad \text{and} \quad r(2) = \frac{k_2}{1+(k_1 + k_1 k_2)^2+k_2^2}~,
\end{equation}
and $r(s) = 0, \forall s > 2, s \in \mathbb{Z}$.
When using autocorrelations as summary features, the SNPE-C procedure learns to invert these relations and obtain the values of $k_1$ and $k_2$ in terms of $r(k)$.
\end{document}